\documentclass{article}

     \PassOptionsToPackage{numbers, compress}{natbib}

\usepackage[preprint]{neurips_2020}

\usepackage[utf8]{inputenc} 
\usepackage[T1]{fontenc}    
\usepackage{hyperref}       
\usepackage{url}            
\usepackage{booktabs}       
\usepackage{amsfonts}       
\usepackage{nicefrac}       
\usepackage{microtype}      

\usepackage[pdftex]{graphicx}
\usepackage[]{color,soul}
\usepackage{textcomp}
\usepackage{enumitem}

\title{Generalization on the Enhancement of Layerwise Relevance Interpretability of Deep Neural Network}

\author{%
  Erico Tjoa\\
  Nanyang Technological University\\
  HealthTech Division, Alibaba Inc\\
  \texttt{ericotjo001@e.ntu.edu.sg} \\
   \And
  Guan Cuntai\\
  Nanyang Technological University\\
  \texttt{ctguan@ntu.edu.sg} \\
}

\begin{document}

\maketitle

\begin{abstract}
The practical application of deep neural networks are still limited by their lack of transparency. One of the efforts to provide explanation for decisions made by artificial intelligence (AI) is the use of saliency or heat maps highlighting relevant regions that contribute significantly to its prediction. A layer-wise amplitude filtering method was previously introduced to improve the quality of heatmaps, performing error corrections by noise-spike suppression.  In this study, we generalize the layerwise error correction by considering any identifiable error and assuming there exists a groundtruth interpretable information. The forms of errors propagated through layerwise relevance methods are studied and we propose a filtering technique for interpretability signal rectification taylored to the trend of signal amplitude of the particular neural network used. Finally, we put forth arguments for the use of groundtruth interpretable information. 
\end{abstract}

\section{Introduction}
The success of deep learning (DL) predictive capability has not been followed by the understanding of its inner working mechanism. Generally, DL algorithms are still considered black-boxes. The need to understand the algorithms becomes clear especially when machine learning (ML) and neural network (NN) algorithms start to permeate into many aspects of the society, such as the medical sector, where accountability and responsibility are of utmost importance. Thus, a variety of attempts to understand them have emerged \citep{CausabilityMedicine, MythosInterpretability, XAIUndisInt, SurveyXAIMed}. Amongst them, visualization method is popular for understanding machine decision. Class Activation Maps (CAM) and its variants \cite{CAM, GradCAM} have been developed to extract more precise features in a data sample that contributes to the prediction. The information is presented as saliency maps or heatmaps, which are often considered readily interpretable and reader-friendly. Likewise, Layerwise Relevance Propagation (LRP) \cite{PixelWiseLRP,CleverHans} produces saliency map by decomposing and back-propagating modified signals that have been fed-forward during the prediction phase. In this manner, the prediction is “explained”, in particular by normalization and selection of positive signals. Also see guided backpropagation \cite{GuidedBP} and the excellent interactive presentation of heatmap visual comparisons in \cite{smoothgrad}. 

The quality and effectiveness of interpretable heatmaps have been demonstrated in several ways. CAM \cite{CAM} and GradCAM \cite{GradCAM} heatmaps were shown to improve the localization on ILSVRC datasets. \cite{DeepLIFT} provides an ingenious method to pick important pixels to the prediction of a class \(c_0\) compared to one other selected class \(c_i\). These pixels are deleted, and the desired change in log odds scores is successfully demonstrated. In \cite{LRPEnch}, 3D U-Nets are used to perform lesion segmentation using multi-modal MRI data, a process of automation aimed at improving stroke diagnosis pipeline. LRP is used to extract interpretable information to analyse the region in the MRI data that contributes to the localization of lesion. Inclusivity coefficient is used to quantify the heatmap quality. 

However, the heatmaps themselves are often not transparently presented. For some of the saliency methods mentioned above, \textit{conceptual uniformity} may not have been attained. We define \textit{conceptual uniformity} as the consistent attribution of relevance to features within or outside the object of interest being classified, segmented etc. Some heatmaps may highlight inexplicably different regions for the same class of prediction, even with correct predictions (consider \cite{PredDiffAnalysis, ACE_NIPS2019_9126,Kindermans2019Unreliability}). Note that the standard for conceptual uniformity is not yet established, but preferably it is flexible enough to accommodate prediction based on the contribution of background information (otherwise the problem becomes similar to instance segmentation). Using the ideas defined in \cite{conf_icml_KimWGCWVS18}, this means the concept activation vectors (CAV) are pointing towards similar directions for a specific concept.

This paper extends the work in \cite{LRPEnch}, which hypothesized that heatmap errors are propagated through spikes in the LRP signals in the U-Net. \cite{LRPEnch} thus applied amplitude filtering to reduce the spikes to rectify the LRP signals. The contributions of this paper are the following:
\begin{enumerate}[leftmargin=*]
\item We present a theoretical framework on the effect of filtering for the rectification of interpretability signals, generalizing amplitude filtering for the LRP signals in \cite{LRPEnch}. The framework assumes that groundtruth interpretable information exists and shows that optimized filter parameters could put an upper bound to the mean error value.
\item We show that the hypothesis regarding error propagation in \cite{LRPEnch} is not general by defining the quantity MP (see later). It is however still possible to apply similar logic to rectify LRP signals through filtering, as suggested by point 1. In particular, if spikes are informative features (rather than errors) in a particular NN, then we use amplitude amplification filters (rather than amplitude clamp filters) for the spikes to improve the quality of interpretable information.
\end{enumerate}

The importance of interpretability groundtruth is not only theoretical, but also evident in practice. Observe the relevance provided by the heatmap shown in figure 1 of \cite{LRP2017}. Number 3 is predicted by supposedly considering the relevance of the regions around the middle protrusion of the digit 3. We can agree with this easily with human intuition. However, without conceptual uniformity (i.e. when there are many other predictions of the digit 3 with different heatmaps) the explanation provided by the heatmap \cite{LRP2017} is said to be given \textit{after the fact} and suffers from \textit{hindsight justification}. If groundtruths for heatmaps are available, we can avoid such arbitrariness. With groundtruth, inexplicable parts of the heatmaps such as label 8 "Military Uniform" (observe the stars in the jet-plane) and label 13 "bison" (the wirenet appears with high attribution values) \cite{smoothgrad} might be improved. Deciding on such groundtruths may also help spur discussions on what is considered interpretable information. As an example, in figure 1 of \cite{CAM}, the heatmap for the classification "cutting-tree" may be reconsidered, perhaps with higher attribution on the whole tree. As for \cite{LRPEnch}, the metric used to measure the quality of heatmaps does not contain constraint from medical knowledge. It acknowledges that a robust relevant medical knowledge is not available, raising possible questions of what groundtruth should be provided. The explainable AI community may need more discussions on what constitutes relevant interpretable information.

\begin{figure*}[hb]
\centering
\includegraphics[width=4.8in]{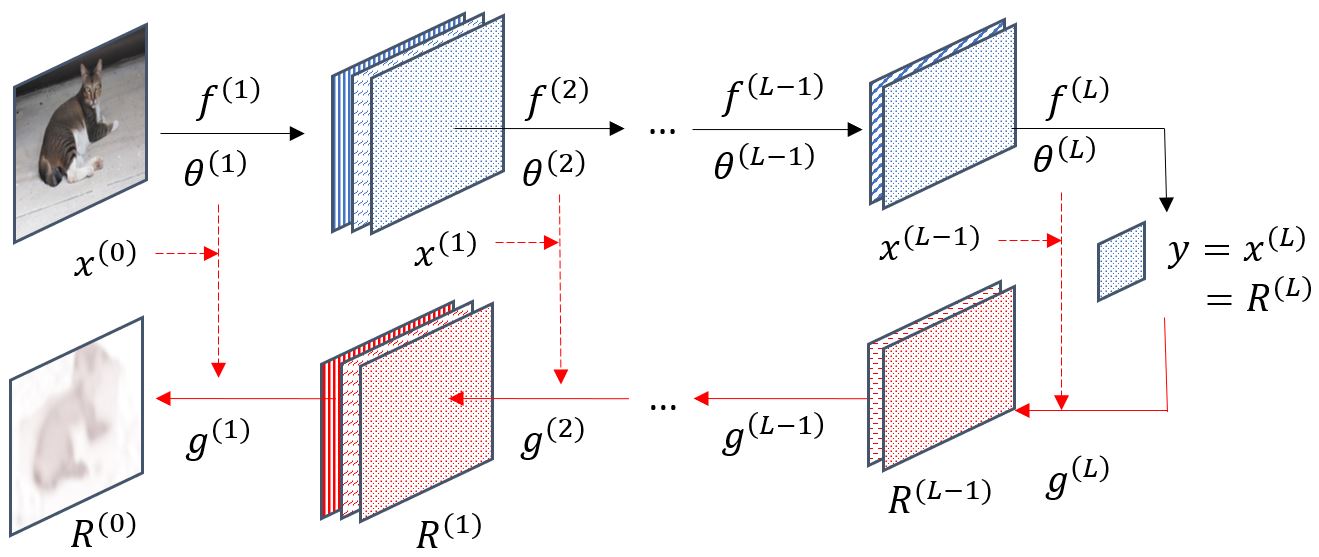}
\caption{Outline of LRP. Feed-forward network (blue) and the relevance propagation (red). Prediction is denoted y, which is also the first relevance \( R^{(L)} \). Relevance \(R^{(k)}\) uses input \(x^{(k)}\), parameters \(\theta^{(k)}\) and the relevance from the layer before it \(R^{(k+1)}\). The final relevance \(R^{(0)}\) is the “explanation” produced by the algorithm. In this case, the classification of the image into "cat" category utilizes the information in \(R^{(0)}\) that highlights regions that resemble the silhouette of the cat, implying that the network is indeed looking at the right place in its decision making.}
\label{fig:LRPIllu}
\end{figure*}

\section{Background and Definitions}
In this section, we aim to generalize the structure of the algorithm used to generate “explanation” used by LRP while retaining its layered form, as shown in figure \ref{fig:LRPIllu}. In this work, \(g(.)\) is used in place of LRP, although it should also admit any different algorithms that work by propagating some ``relevance” values back through the layers with trainable weights.

\textbf{Image-like components}. Each \(x^{(k)},R^{(k)}\) is image-like; the superscript (k) is used here to denote layer k. We can set \(x^{(0)}\in [0,1]^{c\times w \times h}\) where \(c=1\) corresponds to greyscale image, \(c=3\) to the three RGB channels, \(w\) width and \(h\) height. Feature maps are given by \(x^{(k)}\in \mathbb{R}^{3\times w \times h}\), and they could be for example the usual feature maps from convolutional layers or a layer in the dense network. \(y=x^{(L)}=R^{(L)}\) is the predictive output for the particular image \(x^{(0)}\), where we do not include the index denoting the n-th data point in the dataset. 

Since these components are image-like, they have values in spatial domains. Indeed, we can treat \(x^{(k)}\) as functions \(x^{(k)}=x^{(k)}(u)\) and \(u\in U^k\) is for example, discrete index \((c_i,w_i,h_i)\). In this case, \(f^{(k)}\) is a functional. We can also treat \(x^{(k)}\) as a high-dimensional vector instead, which will be more convenient for some computation.

\textbf{Neural network formula and relevance propagation}. We can then write the NN and the relevance propagation (the algorithm to generate “explanation”) in the recursive form
\begin{equation}
x^{(k+1)}(u)=f^{(k+1)}(u,\theta^{(k+1)},x^{(k)}),u\in U^k
\label{eq:forwardprop}
\end{equation}
\begin{equation}
R^{(k)}(u)=g^{(k+1)}(u,R^{(k+1)},\theta^{(k+1)},x^{(k)}),u\in U^k
\label{eq:lrpprop}
\end{equation}
In this work, at each layer, spatial domains for both forward and relevance propagation are the same. In particular, when using RGB images of dimension \((3,w,h)\) normalized to \([0,1]\), we have \(R^{(0)}(U^{(0)})\) with \(U^{(0)}=[0,1]^{3\times w\times h}\). This ensures that each pixel in the image has a relevance value. In figure \ref{fig:LRPIllu}, high relevance corresponds to darker patch in \(R^{(0)}\); the region generally points to the position of the cat. The choice to sum up over all channels of \(R^{(0)}\) seems to be, in practice, arbitrarily made.

\textbf{First-order Errors and Arbitrary Errors}. We are interested in separating “small” errors from “large” errors in the signals that are used to propagate interpretable information \(R^{(k)}\). The ``large" error is separated from ``small" error, inspired by the observation that some erroneous signals are spike-like, very large. How ``small" or ``large" the error is can be relative. However, in this work, we will refer to the usual first-order Taylor approximation error as the small error, and other errors as large. 

For any image-like component \(X\), we now denote small deviation from \(X\) at spatial coordinates \(u\) as $\tilde{X}(u)=X(u)+e_{1}(u)$ where \(e_1\) denotes first order error. We denote large deviation at spatial coordinate \(u\) as \(\delta X(u)=X(u)+\delta e(u)\) where \(\delta e\) denotes a large deviation. In both cases, \(X(u)\) is the true value at coordinate \(u\). We use set theoretic notation naturally, so that the function \(X'\) of a spatial domain \(U\), as a set, includes both the small and large errors \(X'=\tilde{X}\cup\delta X\). As a standalone set, \(\tilde{X}\) will be used to denote small deviation from \(X\) across its domain \(U\). The domains for \(\tilde{X}\) and \(\delta X\) in \(X'\) are disjoint since \(X'\) is still a function, and we will refer to them mostly as \(U_\alpha, U_{\delta\alpha}\) respectively for an \(\alpha\) to be described later. Also, wherever \(\approx\) sign is used, it will correspond to the small error.

\textbf{Contracted Notation}. We first focus on one arbitrary layer and replace equation (\ref{eq:lrpprop}) with \(R_0 (u)=g(u,R_1,X)\) where \(X\) includes both input feature map to \(x\) and parameter \(\theta\). The layer \(R_0(u)\) denotes true relevance since it contains no terms with errors such as \(\tilde{X},X',\tilde{R}\). Note that this generalization is applicable for interpretability methods like LRP. The LRP process itself is independent of optimization that occurs during the training process. It thus takes in both the weights \(\theta\) and image \(X\) both as input on the same footing, justifying our contraction to \(X\).

\section{Denoising Explanation with \(F_\alpha\)}
Suppose we have a sub-optimally trained NN, i.e. an NN not yet achieving the desired evaluation performance. Studying sub-optimal NN may help provide more information about how to better guide the training process of a NN, although we acknowledge the subtle difficulty in deciding how optimal is optimal where evaluation performance such as accuracy cannot yet attain, say, \(99\%\). Nevertheless, let us proceed by denoting a sub-optimally trained layer as \(x'(u)=f(u,\theta',x_1')\). Recall that we use apostrophe symbol ‘ to denote large error. Then, using contracted notation, suppose a relevance signal at this layer is \(R_0'=g(u,X',\tilde{R}_1),u\in U\) i.e. the relevance propagated from the previous layer \(\tilde{R}_1\) is already “denoised”, no longer having large errors (thus no longer denoted \(R_1'\)). The objective here is to apply filter that denoise \(R'_0\) to \(\tilde{R}_0\) so that it can be propagated to the next layer without carrying over large error. See a later section \textit{Unrectified Relevance Propagation} where we observe the consequence of using \(R_1'\) instead.

\textbf{Error separability and \(F_\alpha\)}. Now, let the relevance be \textit{error separable}, \(U=U_{\alpha}\cup U_{\delta\alpha}\) such that \(g(u\in U_\alpha,X',\tilde{R}_1)\approx g(u\in U_{\alpha},\tilde{X},\tilde{R}_1)\equiv g_\alpha\) and \(g(u\in U_{\delta\alpha},X',\tilde{R}_1)\equiv g_{\delta\alpha}\) is still largely erroneous. We have \(g=g_\alpha \cup g_{\delta\alpha}\) separated by small and large error domains and note that this is meaningful only when volume of error \(|U_{\delta\alpha}|\) is manageable (see next section). In another words, there exists a property distinguishing large error encoded in some property \(P=P(g,U_{\delta\alpha},x',X',\tilde{R}_1)\) (see example later).

Then, we define the corresponding denoising function for the separable error as \(F_\alpha\) such that \(F_{\alpha}[g]=g_{\alpha}\cup\hat{F}_\alpha [g_{\delta\alpha}]\) for some \(\hat{F}_\alpha\). In \cite{LRPEnch}, this corresponds to fraction-pass or clamp filters. Let us now define sum of errors
\begin{equation}
\label{eq:sumerror}
SE=\Sigma_{u\in U} |g(u,X,R_1)-g(u,X',\tilde{R_1})|\\ = \Sigma_{u\in U_{\alpha}} e_1(u)+\Sigma_{u\in U_{\delta\alpha}} \delta e(u)
\end{equation}
where \(e_1(u)=|g(u,X,R_1)-g(u,X',\tilde{R}_1)|\approx |\Sigma \frac{\partial g}{\partial x_i} dX_i + \Sigma \frac{\partial g}{\partial R_i} dR_i|\) which contains first order errors in the input to explanations. Furthermore, 
\begin{equation}
\label{eq:delta_e}
\delta e(u)=|g(u,X,R_1)-\hat{F}_\alpha [g(u,X',\tilde{R}_1)]|
\end{equation}

\textbf{Volume of domain containing errors}. From \cite{LRPEnch}, what appears to be spike-like errors motivated the use of filters dependent on signal magnitude. Our results in the experiment section later shows that spikes are not the general forms of errors in sub-optimal NN and thus suggestion in \cite{LRPEnch} possibly applies only to its 3D U-Net. However, a spike or an unnaturally large signal could be easily distinguished using computer algorithm, thus we will use it for illustration. Finding the general form of errors is not within the scope of this paper.s

An example of distinguishable large error is finding the subset \(U_{\delta\alpha}\) of \(U\) with property \(P(g,U_{\delta\alpha},x',X'\tilde{R}_1)\) given by \(|g(U_{\delta\alpha})|\ge\alpha\) for some \(\alpha\). This is the starting point that inspired the use of notation \(F_{\alpha}\). If the relevance propagation signal layer is controlled in the sense that \(|g(u\in U)|<\alpha \ll 1_{g}\) where \(1_g\) is the maximum value in allowed in relevant components the system, then errors could thus be easily identified by their high amplitudes. However, this is not generally the case in existing designs of NN, especially with layers using ReLu or other activation functions that have no finite upper bound.

Suppose only a fraction of the domain manifests as large errors (for example spikes), i.e. \(|U_{\alpha}|=(1-p_\alpha)|U|\) for some small \(p_\alpha\), where \(p_\alpha\) is the probability of finding the errors. Then from equation (\ref{eq:sumerror}), we have \(SE=|U_{\alpha}|\Sigma_{u\in U_{\alpha}} \frac{1}{|U_\alpha|} e_1(u)+|U_{\delta\alpha}|\Sigma_{u\in U_{\delta\alpha}} \frac{1}{|U_{\delta\alpha}|}\delta e(u)\) and thus mean absolute error
\begin{equation}
\label{eq:mae}
MAE=\frac{SE}{|U|}=(1-p_{\alpha})E[l(dx_i,dR_i)]+p_{\alpha}E[\delta e(u)] 
\end{equation}
where \(l(x,y)\) is some linear function because first order term from Taylor expansion is simply a linear term weighted by the gradient of the function.

\textbf{Filters, Effectiveness and Their Optimization}. In this section, we show how the optimization of filter parameter is related to the upper bound of errors, as indicated by figure 5 of \cite{LRPEnch}.

Now suppose we apply no filter, that is, we set \(\hat{F}[X]=X\) identity function. Then, from the second term of equation (\ref{eq:delta_e}), using triangle inequality, we have \(\delta e(u)\le |g(u,X,R_1)|+|g(u,X',\tilde{R}_1)|\) hence \(p_{\alpha}E[\delta e(u)]\le p_{\alpha}(E[|g|]+1_{g})\) where:
\begin{enumerate}[leftmargin=*]
\item The following has been used \(E[|g(u\in U_{\delta\alpha}, X, R_1)|]=E[|g(u\in U_{\alpha}, X, R_1)|]\), which is expected in the case of non-erroneous parameters \(X,R_1\).
\item Error from the first term in MAE is linear w.r.t first order errors, and thus is a typical small perturbation. 
\item The error in \(p_{\alpha}E[|g|]\) term is irrelevant for now, because of the next point.
\item There is no best known upper bound for the term \(p_{\alpha}E[|g(u,X',\tilde{R}_1)|]\) as \(X'\) is still present. The error can be arbitrarily large. We represented it as \(p_{\alpha}1_{g}\) above. If \(1_g\sim E[|g|]\), for example when the signals are capped via sigmoid functions, then we need \(p_\alpha\) to be on the order of \(dx_i, dR_i\) or the error to not be disruptive; this is expected. Otherwise, the order of the errors is significant, which is also expected, since no filter is used. Often, \(1_g\sim E[|g|]\) is not true, for example when activation functions used are ReLu, which has not upper bound. This also means there is no upper bound for the error.
\end{enumerate}

\textit{Optimality of filter parameter}. Now, we apply a type of filter by setting \(\hat{F}[X]=\alpha\) and see the difference between its error upper bound with that above with no filter. For spike-like error, using triangle inequality, similarly we have \(p_\alpha E[\delta e(u)]\le p_\alpha (E[|g|]+\alpha)\) with \(E[|g|]<\alpha\) so that \(p_\alpha \alpha\) might still be significant, and only \(p_\alpha \sim dX_i,dR_i\) gives non-disruptive errors. Increasing \(\alpha\) reduces \(p_\alpha\), though the error might spill over into the constants of the linear function \(l(dX_i,dR_i)\) due to the failure to capture the error because of the continuity towards the spikes. This is because as $|U_\alpha|+|U_{\delta\alpha}|$ is the constant corresponding to total size of elements of $U$ (e.g. the total number of pixels from the input), increasing $\alpha$ may reduce $|U_{\delta\alpha}|$ but increases $|U_\alpha|$ where the error spills into. This error will be insignificant in case of very isolated spikes or high gradient spikes, where adjusting $\alpha$ may not alter much the subset of the domain containing errors. This expresses the optimum \(\alpha\) setting that could be attained depending on the gradient of the spikes, as suggested by figure (5D) of \cite{LRPEnch}. As mentioned earlier in this section, this \(p_{\alpha}\alpha\) is the term allows for optimization of error minimization.

Finally, if \(\hat{F}[X]=0\), we are left with \(p_{\alpha}E[|g|]\). This provides more relaxed condition to achieve small error compared to \(\hat{F}[X]=\alpha >E[|g|]\) because both \(p_\alpha\) and \(E[|g|]\) can contribute to the error upper bound of the order of \(dX_i,dR_i\). Expecting \(E[|g|]\) to be on the order of \(dX_i,dR_i\) means flattening \(|g|\) to the level of uniform function with local variance, which might be feasible in the case of images with rather uniform datasets. Otherwise, in natural images, it might mean changing color channels and reducing color contrasts. In any case, optimizing \(p_\alpha\) and choosing optimum degree of “flattening” could give a low error upper bound as well.

\textbf{Unrectified Relevance Propagation}. Now, we observe the consequence of using \(R_1'\) instead, i.e. we attempt to rectify signals given unrectified signal. Suppose \(U=A\cup B\) such that \(A\) contains pixels that suffer only small errors even with \(R_1'\), while \(B\) contain the rest. By comparison, in the previous sections, we immediately separate to \(U_{\alpha}\cup U_{\delta\alpha}\). Then \(g(u,X',R')\approx g(u\in A,X',\tilde{R})\cup g(u\in B,X',R'))\). Thus
\begin{equation}
\hat{F}_{\alpha}[g(u,X',R')]=g(u\in A_{\alpha},X',\tilde{R})\cup \hat{F}_{\alpha}[g(u\in A_{\delta\alpha},...)]
\end{equation}
where \(A_{\alpha}\subseteq U_{\alpha}\) because \(A\subseteq U\). Hence, \(|A_{\delta\alpha}\cup B|=|(U_\alpha -A_\alpha)\cup U_{\delta\alpha}|\ge |U_\delta|\) thus \(p_\alpha =|A_{\delta\alpha}\cup B|/|V|\) is a larger error by the amount related to the size of \(U_\alpha - A_\alpha\). This reflects the need to control the error at each layer.

\begin{figure*}[ht]
\centering
\includegraphics[width=5.6in]{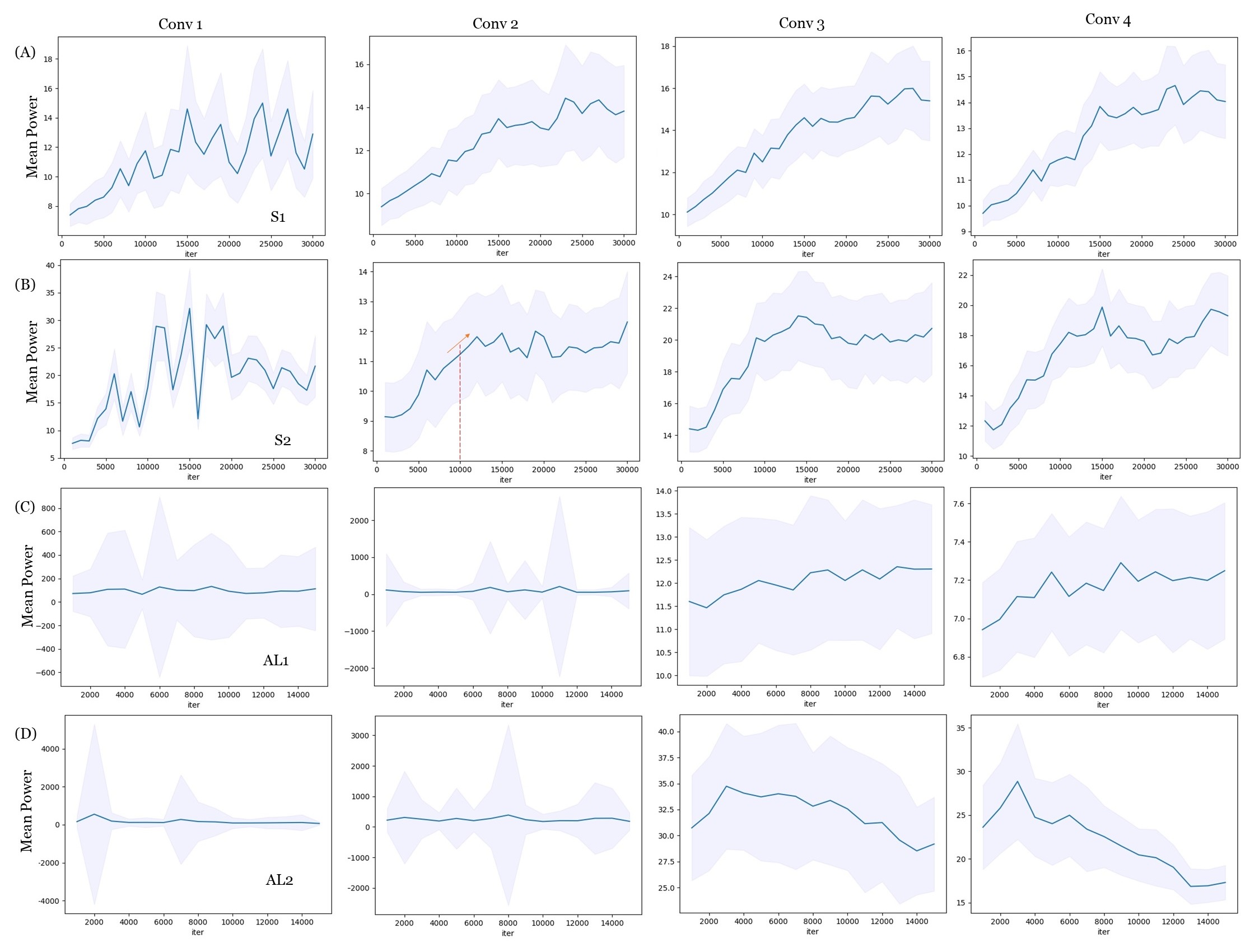}
\caption{Plots of MP values of the first four convolutional layers through the training phase of the neural networks respectively labelled (A) S1 (B) S2 (C) AL1 and (D) AL2. Different scale of input and structure of neural networks yield variations in MPs, as opposed to the suggestion from \cite{LRPEnch} that MPs decrease with the increase in NN performance. Evaluation is performed sparsely (at every 1000 iterations).}
\label{fig:mptrend}
\end{figure*}

\section{Experiment}
\subsection{Experimental Setup}
The neural networks for our discussion are named S1, S2, AL1 and AL2, described in the following and the dataset used is MNIST. First, S1 is a generic, relatively small convolutional neural network (CNN) with 270578 parameters, which we just conveniently call SmallNet. AL1 is like AlexNet with 2664138 parameters. S2 is like S1, except that images are first enlarged by interpolation to \((140,140)\) and the SmallNet is modified so that the no. of parameters is similar, 273381. Likewise, AL2 is similar to AL1 with the same image enlargement and network modification, 2874810 parameters. Adam optimizer is used and batch size \(n_b=4\). Training dataset consists of 60000 images, hence each training epoch consists of \(60000/n_b=15000\) iterations. Each evaluation of accuracy is performed on 1000 handwritten digits randomly drawn from the test dataset, while each evaluation of mean powers (defined in the next paragraph) on 240 randomly drawn from test dataset. To generate heatmaps for interpretability, we use LRP \cite{PixelWiseLRP,CleverHans}. Other settings can be found in the source code\footnote{\url{https://github.com/etjoa003/explainable_ai/tree/master/fcalc}}.

As we are initially interested in observing the spikiness of the errors, we defined the quantity Mean Power, MP. Given threshold \(t\) and a layer signal \(R^{(k)}\), let \(R^+\) be positive part of \(R^{(k)}\) i.e. negative signals are set to zero; likewise \(R^-\) is the absolute value of the negative part of \(R^{(k)}\). Define the upper strength to be \(U_t(X)=mean\{X_i: X_i\ge t\}\) and \(X_i\) is a pixel in \(X\), i.e. the mean value of all the signals in \(X\) (assumed non-negative) that are greater equal to \(t\). Also, define the lower strength to be \(L_t(X)=mean\{X_i:X_i<t\}\). Then \(MP^{+}=\frac{U_t(R^+)}{L_t(R^+)}\), \(MP^-=\frac{U_t(R^-)}{L_t(R^-)}\) and \(MP=\frac{1}{2}(MP^+ +MP^-)\). Spikier signals will give larger \(MP\), and both positive and negative MPs are not sensitive to the absolute amplitude of the signal.

Following \cite{LRPEnch}, clamp-filters and fraction-pass filters with \(\alpha=0.05,0.2,0.6\) are applied during layerwise backward propagation. However, after observing the trend, we apply also fractional-amplifier with \(\alpha=0.5,0.7\) at some of the layers. For relevance signal \(R\), it performs conditional pixel-wise signal amplification \(f_{\alpha,A}(R_i)=AR_i\) if \(\frac{|R_i|}{R_{max}}\ge\alpha\) and no effect to the pixel otherwise, where \(R_{max}\) is the maximum absolute value across \(R\)'s pixels, and subscript \(i\) is an index of some pixel. In this experiment, \(A=2\).

\subsection{Results and Discussions}
{\cite{LRPEnch}} proposes a general hypothesis on the forms of errors being propagated. If we describe it using MP, the proposed hypothesis can be simply summarized as ``MP trend generally decreases". However, the hypothesis may only be applicable to the neural network used in {\cite{LRPEnch}} , i.e. it may not be general. To show that it is not general, we simply need to provide a counter-example on any dataset and model, i.e. by showing that the MP trend is not always increasing. Here, we choose to work with small or standard neural networks and show that some of them do serve as counter-examples. Small neural networks on simple MNIST data are chosen for fair comparison, as each neural networks will achieve equally high accuracy. Indeed, AlexNets AL1 and AL2 quickly exceeded 0.9 accuracy (within around 1000 iterations) while SmallNets S1 and S2 reached it at the end of epoch 2. 

MP trends have been defined to test the way errors are propagated. The hypothesis suggested in \cite{LRPEnch} can be reformulated in terms of MP values: if errors consistently take the form of spikes, then MP values will decrease as neural networks undergo training, assuming that errors are reduced along the training. Hence, to show that the hypothesis is not general, we only need to show some examples of neural networks with MP values that are not decreasing. From figure \ref{fig:mptrend}, the variations in MP trends are clear. S1 and S2 generally have increasing MP values while AL2 shows decrease in MP and AL1 shows a less pronounced increase. On a side note, the same networks with different initialization show some resemblance in their MP trends (not shown), though it is not clear to what extent this applies. 

\begin{figure}[!ht]
\centering
\includegraphics[width=4.2in]{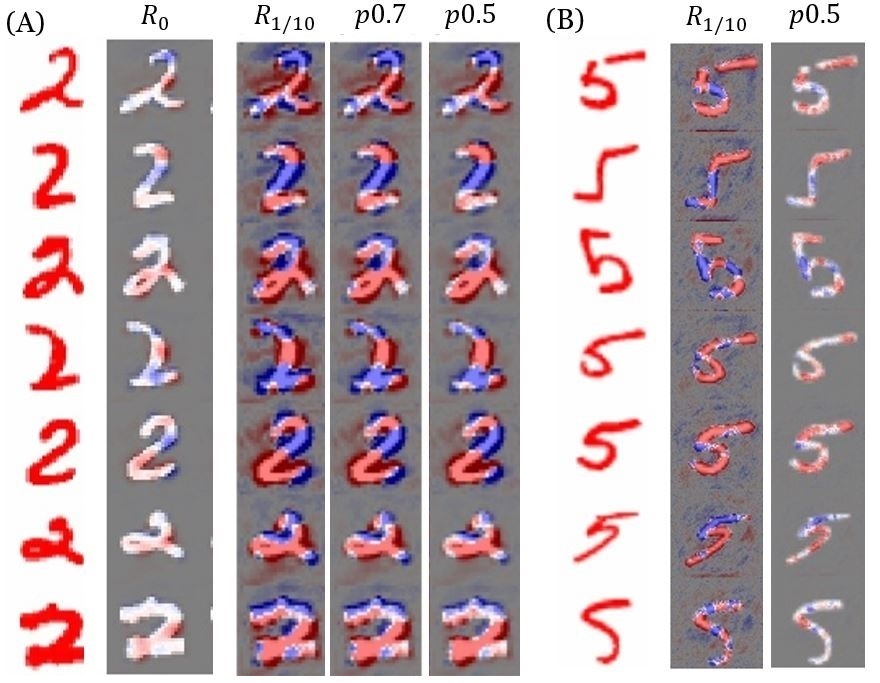}
\caption{\(R_0\) shows the relevance heatmap, corresponding to \(R^{(0)}\) in the earlier definition, normalized to \([-1,1]\). Redder regions are interpreted as important regions to the choice of the prediction, while bluer regions are contributing against the choice of the prediction. \(R_{1/10}\) is \(R_0\) viewed at 1/10 maximum intensity. \(p0.7\), \(p0.5\) (also viewed at \(R_{1/10}\)) are the LRP output after the application of fractional-amplifier with \(\alpha=0.7,0.5\) respectively. This "rectification" appears to reduce noise signals in regions outside the handwritten digits, particularly from (B). (A) is drawn from iteration 30000 of S1 training. (B) is from iteration 10000 of S2 training when the MP values are still showing increasing trend. Please bear in mind it is not our objective to show with this figure that there is actually quantifiable improvement. Regardless, it does demonstrate what we call the lack of conceptual uniformity.}
\label{fig:pfilter}
\end{figure}

In \cite{LRPEnch}, when the errors are perceived to be spikes, culling the spikes with clamp or pass filters appear to "rectify" the heatmaps to some extent. On the other hand, the increasing MP of S1 and S2 shows that for these NNs, the amount of spikes increase with the performance i.e. spikes are instead used to store useful information, and thus we conversely argue that sub-optimality manifests in the form of less-defined hills in the relevance signals, or the lack of spikes. The suggestion from \cite{LRPEnch} that errors in interpretable information are propagated as spikes in relevance signals does not hold; the spiking error may be unique to the 3D U-Net that is used. However, reversing the logic, the "rectification" comes in the form of amplifying the hills in the relevance signals to form spikes, which is why fractional-amplifier is used for S1 and S2. In particular, for S2, amplification result is displayed at iteration 10000, not at the end of the training. At this point, the MP values are still increasing, as marked in figure \ref{fig:mptrend}(B) Conv 2 orange dashed line. Figure \ref{fig:pfilter}(A,B) show some degree of success: if better interpretability means better hot/cold regions relevant to the areas in the images that contain the handwritten digits, then the improvement due to the amplifier with \(\alpha=0.5,0.7\) (shown as \(p0.5, p0.7\) in the figure respectively) is to some degree shown by lower intensity noises outside the digits. In this argument, it is assumed that the groundtruth exactly overlaps with non-zero-intensity region of the digits (in another words, the digit itself). So far, we have addressed objective (2) of the paper without quantifying what it means to be ``better" heatmaps. In fact, here, the ``improvement" is only dependent on our visual observation, where heatmaps are closer to the assumed segmentation mask than the noisy version. This brings us back to the main point: we need a groundtruth at some point in the future. Before proceeding with it, we explore the notion of ``better" or useful heatmaps with the following short detour.

To illustrate the importance of conceptually uniform groundtruth for interpretable information, we raise a small question on the utility of heatmaps. First, observe figure {\ref{fig:pfilter}}. The prediction for number ``2" does not yield heatmaps of the same shapes. Next, observe figure 1 in the supp. material. We give a hindsight justification on the digit 2 boxed in green. Higher intensity red region contributes to the prediction ``2", particularly the open curve in the center and left regions. Blue region is the opposite; it might indicate ``7", although the tail region (bottom-right) helps conclude the prediction as ``2". With such interpretation, the question is, which of the digit(s) in figure 1 of supp. material is/are correctly predicted? Is it possible to predict them reliably? Such situations lack conceptual uniformity, and the main suggestion of current work is precisely to avoid this by showing the importance of providing groundtruth heatmaps.

Back to the main point, if groundtruths are available, we might be able to utilise demonstrate quantitatively improvements in interpretability, for example, using equation \ref{eq:mae}. Consider figure \ref{fig:pfilter}(A) first row. With estimates (see supplementary materials \textit{error estimates}), \(MAE\approx 0.03, 0.015\) for \(p0.7\) and \(p0.5\) respectively. Finally, the answer to the previous question is, all predictions for number 2 and 1 in the figure 1 of supp. material are correct. It is correct even for the last row, where only a few local high intensities (spikes of intensities) contribute to the prediction. High relevance regions are hard to pinpoint, and this might raise doubts to the usefulness of heatmaps. For completeness sake, in figure \ref{fig:pfilter}(A,B), the predictions of 2s and 5s are 1101101 and 1100000 going down the row where 1 is correct and 0 is wrong. A heatmap has many potentials as the representation of interpretable information. It might need some improvements to boost its credibility, and providing groundtruth for interpretable information may be a solution. Training NN architecture that optimizes against the correct, desired groundtruth might be helpful, though admittedly the process may incur more resources required for manual annotation and labelling.

\section{Conclusion} 
This paper presents two main results: (1) theoretical framework for the use of filters to put upper-bounds on noise-induced errors when groundtruth interpretability information exists and (2) an experimental demonstration to show the variety of errors propagated through layerwise relevance methods. Although general trend of spike-like errors suggested in \cite{LRPEnch} is not observed, the technique of applying filters to counter the trend of MP values can possibly be used to reduce the errors of the heatmaps. The results have been presented in the hope of spurring more research into the direction of having robust, quantifiable and verifiable interpretable information.

\section*{Broader Impact}
Providing interpretable information for the automation of some tasks, such as medical diagnosis, is very important. Due to the lack of transparency in the working mechanism of a neural network, wrong prediction or misdiagnosis lower the credibility of a neural network, hence preventing its practical deployment. However, while interpretability research has been booming, the presentation of improvements in interpreteable information is beset by the lack of agreeable framework that match human understanding. Either a critical assessment of displayed heatmap is presented or a completely different pipeline on how interpretable information is displayed should be developed. 

\bibliographystyle{unsrt}

\bibliography{fcalctnips}

\section{Supplementary Materials}

\section{Error Estimates}
Consider figure 3(A) in the main text. If we take the groundtruth as the \(p0.5\) minus all the hot/cold region outside the digit, then the error in \(p0.7\) is approximately the following. For \(\alpha=0.7\) let \(p_{\alpha}\approx 0.3\) by visual observation, i.e about 1/3 erroneous region. Errors can be considered perturbative since the heatmaps core hot/cold regions between \(R_{1/10}\), \(p0.7\) and \(p0.5\) are overlapping. The heatmap is normalized, i.e. the order is 1. The view is 1/10, the linear part of error can be considered like the usual perturbative error, assume of the order 0.001. Thus the first term is \((1-0.3)\times\frac{1}{10}\times 0.001=7\times 10^{-5}\). The \(\delta e(u)\) between \(p0.7\) and \(p0.5\) (which we assume to be groundtruth-like) appears to be approximately half-intensity of the erroneous region in 1/10-view, giving the second term \(0.3\times\frac{1}{10}=0.03\) which dominates the error. Hence, \(MAE\approx 0.03\). Likewise, the \(p0.5\) error is dominated by the second term, although the erroneous region appears in half the intensity, thus \(MAE\approx 0.15\). In another words, \(p0.5\) gives smaller error. Using similar methods to compute the error for intermediate \(\alpha\) values, we may arrive at figure 2 of supp. material that shows an apparent trend of improved heatmap quality. We do not include the full plot in the main text, since the analysis assumes that the parts of the digit are the only correct contributor to interpretable information (i.e. ignoring all other background information, even if it is blank in the case of MNIST dataset). In reality, this is akin to ignoring background of an image, which may actually be informative.

\begin{figure*}[htbp!]
\centering
\includegraphics[width=3.2in]{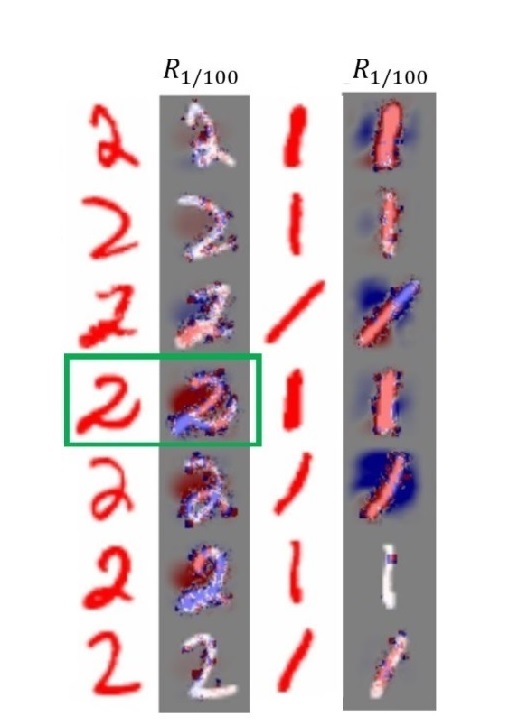}
\caption{Which of the above handwritten digits are correctly predicted?
The number 2 boxed in green and its heatmap are used for hindsight reasoning for reader to answer the question in the main text. Heatmaps are generated
from AlexNets AL2 using LRP.}
\end{figure*}

\begin{figure*}[htbp!]
\centering
\includegraphics[width=3.2in]{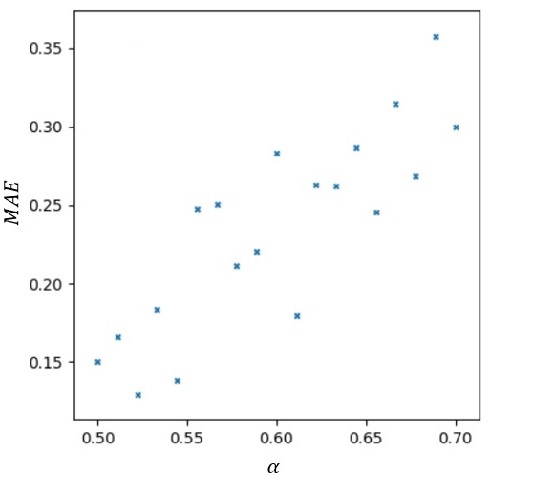}
\caption{Interpolated Error Approximation. Hypothetical representation of interpretable information improvements. This is to illustrate the post-processed
quantification of improvement in interpretable information, assuming that only the digits contribute to interpretable information.}
\label{fig:mptrend}
\end{figure*}

\end{document}